\documentclass[letterpaper, 10 pt, conference]{ieeeconf}  

\IEEEoverridecommandlockouts                              

\title{\LARGE \bf
DeepHybrid: Deep Learning on Automotive Radar Spectra and Reflections for Object Classification
}

\author{Adriana-Eliza Cozma$^1$, Lisa Morgan$^2$, Martin Stolz$^3$, David Stoeckel$^1$, Kilian Rambach$^1$
\thanks{$^1$Bosch Center for Artificial Intelligence, Germany, $^2$Bosch Engineering GmbH, Germany,	$^3$Robert Bosch GmbH, Germany}%
\thanks{$^1${\tt\small kilian.rambach@de.bosch.com}}%
}

\usepackage{graphicx}
\graphicspath{{figures/}}
\usepackage{xcolor}
\usepackage{amsmath}

\usepackage[caption=false,font=footnotesize]{subfig}

\usepackage{stfloats}

\usepackage{siunitx}

\usepackage{slashbox}

\newcommand{\range}{r}
\newcommand{\vel}{v}
\newcommand{\azi}{\vartheta}

\newcommand{\carrierfreq}{f_c}
\newcommand{\bandwidth}{B}
\newcommand{\meastime}{T_{meas}}
\newcommand{\dB}{\si{\decibel}}
\newcommand{\figref}[1]{Fig.~\ref{#1}}

\newcommand{\secref}[1]{Sec.~\ref{#1}}
\newcommand{\cf}{cf.\ }
\newcommand{\ie}{i.\,e.\ }
\newcommand{\eg}{e.\,g.\ }
\newcommand{\wrt}{w.\,r.\,t.\ }

\begin{document}

\maketitle
\thispagestyle{empty}
\pagestyle{empty}

\begin{abstract}
Automated vehicles need to detect and classify objects and traffic participants accurately.
Reliable object classification using automotive radar sensors has proved to be challenging.
We propose a method that combines classical radar signal processing and Deep Learning algorithms. 
The range-azimuth information on the radar reflection level is used to extract a sparse region of interest from the range-Doppler spectrum.
This is used as input to a neural network (NN) that classifies different types of stationary and moving objects.
We present a hybrid model (DeepHybrid) that receives both radar spectra and reflection attributes as inputs, \eg radar cross-section. Experiments show that this improves the classification performance compared to models using only spectra.
Moreover, a neural architecture search (NAS) algorithm is applied to find a resource-efficient and high-performing NN.
NAS yields an almost one order of magnitude smaller NN than the manually-designed one while preserving the accuracy.
The proposed method can be used for example to improve automatic emergency braking or collision avoidance systems.
\end{abstract}

\section{Introduction}
Automated vehicles require an accurate understanding of a scene in order to identify other road users and take correct actions.
Besides precise detection and localization of objects, a reliable classification of the object types in real time is important in order to avoid unnecessary, evasive, or automatic emergency braking maneuvers for harmless objects. 
Typically, camera, lidar, and radar sensors are used in automotive applications to gather information about the surrounding environment. Here we consider radar sensors, which are robust to difficult lighting and weather conditions, and are used as stand-alone or complementary sensors to cameras \cite{mukhtar2015vehicle}.

Up to now, it is not clear how to best combine classical radar signal processing approaches with Deep Learning (DL) algorithms.
Here, we use signal processing techniques for tasks where good signal models exist (radar detection) and apply DL methods where good models are missing (object classification).
We propose a method that detects radar reflections using a constant false alarm rate detector (CFAR) \cite{rohlingCFAR}, and associates the detected reflections to objects. 
This information is used to extract only the part of the radar spectrum that corresponds to the object to be classified, which is fed to the neural network (NN). 
In this way, the NN has to classify the objects only, and does not have to learn the radar detection as well.
In order to associate reflections to objects, the angles (directions of arrival (DOA)) of the reflections have to be determined.
Compared to methods where the angular spectrum is computed for all range-Doppler bins, our method requires lower computational effort, since the angles are estimated only for the detected reflections.
Our approach works on both stationary and moving objects, which usually occur in automotive scenarios.
A range-Doppler-like spectrum is used to include the micro-Doppler information of moving objects, and the geometrical information is considered during association.
Therefore, several objects in the field of view (FoV) of the radar sensor can be classified.
To improve the classification accuracy, we use a hybrid approach and input both radar reflection attributes, \eg the radar cross-section (RCS), and radar spectra into the NN. We call this model DeepHybrid.
We show that additionally using the RCS information as input significantly boosts the performance compared to using spectra only.
Manually finding a high-performing NN architecture that is also resource-efficient \wrt an embedded device is tedious, especially for a new type of dataset.
Therefore, we deploy a neural architecture search (NAS) algorithm to automatically find such a NN.

Related approaches for object classification can be grouped based on the type of radar input data used.
Radar-reflection-based methods first identify radar reflections using a detector, e.g. CFAR \cite{rohlingCFAR}. Then, different attributes of the reflections are computed, \eg range, Doppler velocity, azimuth angle, and RCS. After applying an optional clustering algorithm to aggregate all reflections belonging to one object, different features are calculated based on the reflection attributes. These are used by the classifier to determine the object type \cite{schubert2015-dbscan, Scheiner_2019, prophet2018_car_dog_ped_classifier}.
Compared to radar reflections, using the radar spectra can be beneficial, as no information is lost in the processing steps.
The authors of \cite{rohling-car-pedestrian-features-2010, bartsch2012-pedestrian-recognition} take the radar spectrum into account to compute additional features for the classification, and \cite{prophet-image-based-pedestrian-classification2018} uses feature extractors known from vision to apply them on the radar spectrum. 

In contrast to these works, data-driven DL approaches learn a rich representation in an end-to-end training, such that no additional feature extraction is necessary. DL methods have been very successful in other domains, \eg vision or audio \cite{lecun2015deep}. 
The authors of \cite{schumann2018SemanticSegmentationRadar} apply DL methods directly on the radar reflections. 
In \cite{ensembleLombacher2017,staticObjectClassificationDL-lombacher2016} an occupancy grid based on radar reflections is computed, on which a convolutional neural network (CNN) is applied. 
However, a long integration time is needed to generate the occupancy grid. Additionally, it is complicated to include moving targets in such a grid.
In \cite{convLSTM-Radar-microDoppler2019} the range-Doppler spectrum is computed for multiple cycles, and a combination of a CNN and Long-Short-Term-Memory (LSTM) neural network is used for a 2-class classification problem. However, only 1 moving object in the radar sensor's FoV is considered, and no angular information is used. The range-azimuth spectra are used by a CNN to classify different kinds of stationary targets in \cite{deepLearningClassificationRadarPatel2019}.
A hybrid approach combining radar reflections and spectra is investigated in \cite{3dRadarcube-Cnn-classification2020}. Each reflection is first classified individually and then clustered. Only moving targets are considered, which leads in general to more reflections, since they can be resolved in Doppler domain as well. This makes the classification of stationary targets more difficult. We consider both stationary and moving targets and first associate the reflections before classifying the object.

Compared to these related works, our method is characterized by the following aspects:
1) We combine signal processing techniques with DL algorithms. For each object, a sparse region of interest (ROI) is extracted from the range-Doppler spectrum, which is used as input to the NN classifier.
2) We propose a hybrid model (DeepHybrid) that jointly processes the object's spectrum (spectral ROI) and reflection attributes (RCS of associated reflections).
3) The NN predicts the object class using only the radar data of one coherent processing interval (one cycle), \ie it is a single-frame classifier.
4) The reflection-to-object association scheme can cope with several objects in the radar sensor's FoV. Since part of the range-Doppler spectrum is used, both stationary and moving targets can be classified.
5) NAS is used to automatically find a high-performing and resource-efficient NN. To the best of our knowledge, this is the first time NAS is deployed in the context of a radar classification task.
	
\section{Measurement Setup and Data Preprocessing}
In the following we describe the measurement acquisition process and the data preprocessing.
Typical traffic scenarios are set up and recorded with an automotive radar sensor. The obtained measurements are then processed and prepared for the DL algorithm.

\subsection{Radar Sensor}
To record the measurements, an automotive prototype radar sensor with carrier frequency $\carrierfreq = \SI{76.5}{GHz}$, bandwidth $\bandwidth = \SI{850}{MHz}$, and a coherent processing interval $\meastime = \SI{16}{ms}$ is deployed.
It uses a chirp sequence-like modulation, with the difference that not all chirps are equal. Each chirp is shifted in frequency \wrt to the former chirp, \cf \cite{JFMCW-Patent-Schoor2018} and \cite{radar-waveform-kronauge2014-journal} for a related modulation. This modulation offers a reduction of hardware requirements compared to a full chirp sequence modulation by using lower data rates and having a lower computational effort. 
The range $\range$ and Doppler velocity $\vel$ are not determined separately, but rather by a function of $\range$ and $\vel$ obtained in two dimensions, denoted by $k, l = f(\range,\vel)$. Generation of the $k,l$-spectra is done by performing a two dimensional fast Fourier transformation over samples and chirps, \ie fast- and slow-time.
Note that our proposed preprocessing algorithm, described in \ref{sec:data_preprocessing}, is not specific to this kind of modulation, but also works for radar sensors using a standard chirp sequence modulation.

\subsection{Measurements}
\begin{figure}[b]
	\centering
	\subfloat[pedestrian dummy]{\includegraphics[width=0.33\linewidth, keepaspectratio]{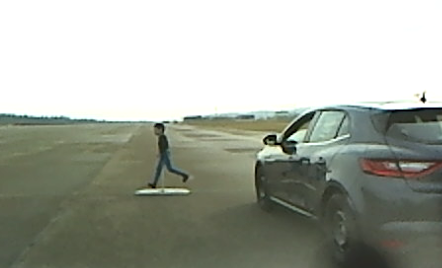}}
	\hfill
	\subfloat[bicycle dummy]{\includegraphics[width=0.33\linewidth, keepaspectratio]{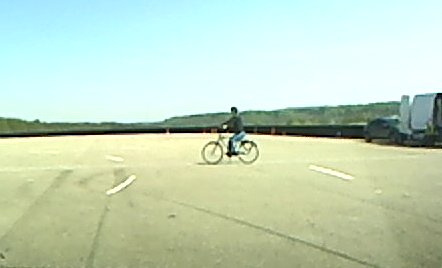}}
	\hfill
	\subfloat[object that can be run over next to parked car]{\includegraphics[width=0.33\linewidth, keepaspectratio]{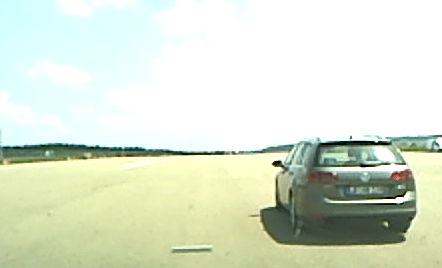}}
	\caption{Measurement examples of different object types} 
	\label{fig_measurement_examples}
\end{figure}

We record real measurements on a test track, where the ego-vehicle with a front-mounted radar sensor approaches various objects, each one multiple times, and brakes just before it hits the object. 
We consider 8 different types of parked cars, moving pedestrian dummies, moving bicycle dummies, and several metallic objects that lie on the ground and are small enough to be run over, see \figref{fig_measurement_examples}.
The metallic objects are a coke can, corner reflectors, and different metal sections that are short enough to fit between the wheels.
The objects are grouped in 4 classes, namely \textbf{car}, \textbf{pedestrian}, \textbf{two-wheeler}, and \textbf{overridable}. 
The pedestrian and two-wheeler dummies move laterally \wrt the ego-vehicle. Therefore, the observed micro-Doppler effect is limited compared to a longitudinally moving pedestrian, which makes it harder to classify the laterally moving dummies correctly \cite{bartsch2012-pedestrian-recognition}.

The measurement scenarios should cover typical road traffic situations, as described by Euro NCAP, for more details see \cite{NCAP-vru2020,NCAP-CAR-protocol2020}.
They can also be used to evaluate the automatic emergency braking function. If there is a large object, \eg a pedestrian, appearing in front of the ego-vehicle, it should detect and classify the object correctly and brake automatically until it comes to a standstill. On the other hand, if there is a small object that can be run over, \eg a can of coke, the ego-vehicle should classify it correctly and just ignore it.

After the objects are detected and tracked (see \secref{sec:data_preprocessing}), the object tracks are labeled with the corresponding class. These labels are used in the supervised training of the NN.
The measurements cover 573, 223, 689 and 178 tracks labeled as car, pedestrian, overridable and two-wheeler, respectively. Each track consists of several frames. 
One frame corresponds to one coherent processing interval.

\subsection{Data Splitting}
We split the available measurements into 70\% training, 10\% validation and 20\% test data. 
The splitting strategy ensures that the proportions of traffic scenarios are approximately the same in each set. 
Thus, we achieve a similar data distribution in the 3 sets. 
Note that there is no intra-measurement splitting, \ie all frames from one measurement are either in train, validation, or test set.
Since a single-frame classifier is considered, the spectrum of each radar frame is a potential input to the NN, \ie a data sample. 
There are approximately $45$k, $7$k, and $13$k samples in the training, validation and test set, respectively.

\subsection{Data Preprocessing}
\label{sec:data_preprocessing}
\begin{figure}
	\centering
	\includegraphics[width=0.99\linewidth, keepaspectratio, angle=0]{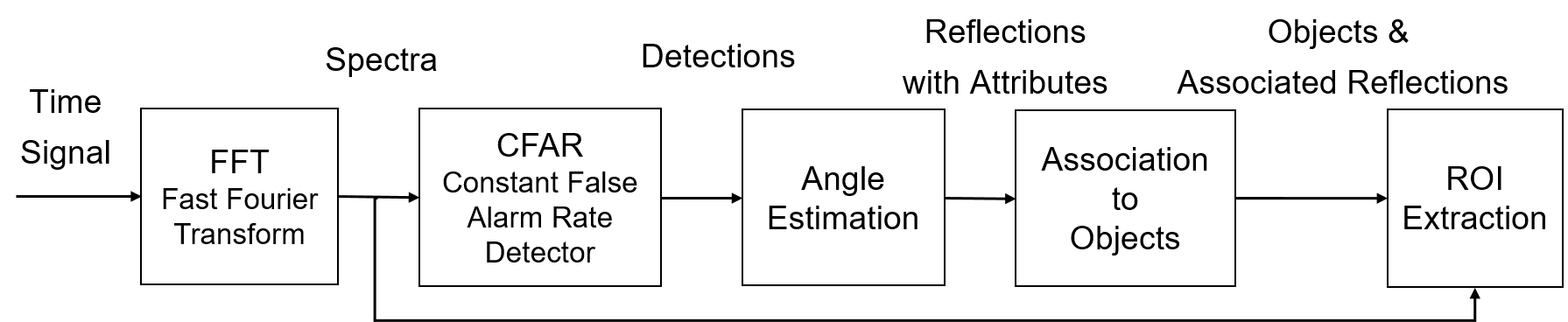}
	\caption{Processing pipeline for the region of interest (ROI) extraction of the spectra. The time signal is transformed into the radar spectrum. After detecting the radar reflections, the angles are estimated, leading to a reflection list with attributes, \ie range $r$, Doppler velocity $v$, azimuth angle $\azi$, and radar cross-section (RCS). 
	The reflections are associated to the objects. Using the $k,l$- or $r,v$- bins of the associated reflections, the corresponding part of the radar spectrum is extracted for each object, yielding the ROI.}
	\label{fig_processing_pipeline}
\end{figure}

The processing pipeline from the radar time signal to the part of the radar spectrum that is used as input to the NN is depicted in \figref{fig_processing_pipeline}. 
The goal is to extract the spectrum's region of interest (ROI) that corresponds to the object to be classified.

First, the time signal is transformed by a 2D-Fast-Fourier transformation over the fast- and slow-time dimension, resulting in the $k,l$-spectra.
Then, the radar reflections are detected using an ordered statistics CFAR detector.
For each reflection, the azimuth angle is computed using an angle estimation algorithm. This results in a reflection list, where each reflection has several attributes, including the range $\range$, relative radial velocity $\vel$, azimuth angle $\azi$, and radar cross-section (RCS).
The RCS is computed by taking the signal strength of the detected reflection and correcting it by the range-dependent dampening and the two-way antenna gain in the azimuth direction.
The polar coordinates $\range, \azi$ are transformed to Cartesian coordinates $x,y$. 
These are used for the reflection-to-object association.

\begin{figure}[t]
	\centering
	\includegraphics[width=0.98\linewidth, keepaspectratio]{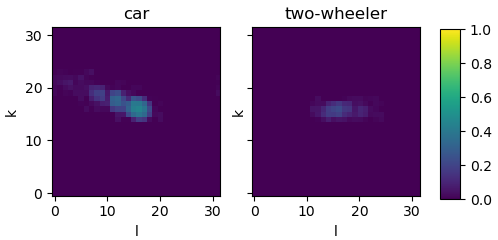}
	\caption{Example of regions of interest (ROIs) for a car and two-wheeler dummy. The axes are relative to the clipped region of the spectrum.}
	\label{fig_extracted_spectra_example_k_l}
\end{figure}

For each associated reflection, a rectangular patch is cut out in the $k,l$-spectra around its corresponding $k$ and $l$ bin. All patches are put together to yield the ROI, which contains only the spectral part of the reflections associated to the object under consideration. The ROI is centered around the maximum peak of the associated reflections and clipped to $32\times32$ bins, which usually includes all associated patches. 
Then, the ROI is converted to $\dB$, clipped to the dynamic range of the sensor, and finally scaled to $[0,1]$. The scaling allows for an easier training of the NN. 
Two examples of the extracted ROI are depicted in \figref{fig_extracted_spectra_example_k_l}. In general, the ROI is relatively sparse.
The object's ROI and optionally the attributes of its associated radar reflections are used as input to the NN.

Here, we focus on the classification task and not on the association problem itself, \ie the assignment of different reflections to one object.
Therefore, we use a simple gating algorithm for the association, which is sufficient for the considered measurements. The approach can be extended to more sophisticated association algorithms, \eg DBSCAN \cite{schubert2015-dbscan}, or methods that take into account the measurement uncertainties in the different dimensions, \eg the Mahalanobis or the association log-likelihood distance \cite{altendorfer_association2016}.

Our proposed approach works with several objects in the FoV of the radar sensor, and can still utilize the radar spectrum, since the spectral ROI for each object is determined. This is important for automotive applications, where many objects are measured at once. Moreover, we can use the $k,l$- or $\range,\vel$-spectra for classification, but still use the azimuth information in addition for association. This is crucial, since associating reflections to objects using only $\range, \vel$ might not be sufficient, as the spatial information is incomplete due to the missing angles. Compared to methods where the complete angular spectrum is computed for all bins in the $\range,\vel$-spectrum, we need to estimate the angle only for the detected reflections, which is computationally cheaper.

\section{Method}
To solve the 4-class classification task, DL methods are applied.
First, we manually design a CNN that receives only radar spectra as input (\textit{spectrum branch}). 
Then, it is shown that this manual design process can be replaced by a neural architecture search (NAS) algorithm, which finds a CNN with similar accuracy, but with even less parameters.
After that, we attach to the automatically-found CNN a sequence of layers that process reflection-level input information (\textit{reflection branch}), obtaining thus the hybrid model we propose.

\subsection{Spectrum Branch}
\subsubsection{Manual Design}
\label{sec:manual}
Several design iterations, \ie trying out different architectural choices, \eg increasing the convolutional kernel size, doubling the number of filters, yield the CNN shown in \figref{fig:arch} (a). 
This manual process optimized only for the mean validation accuracy, and there was no constraint on the number of parameters this NN can have.
The mean validation accuracy over the 4 classes is $\mathcal{A} = \frac{1}{C} \sum_{c=1}^C \frac{p_{c}}{N_c}$
with $C$ being the number of classes, $p_c$ the number of correctly classified samples, and $N_c$ the number of samples belonging to class $c$.
The NN receives a spectral input of shape $(32, 32, 1)$, with the numbers corresponding to the bins in $k$ dimension, in $l$ dimension, and to the number of input channels, respectively. This manually-found NN achieves $84.6\%$ mean validation accuracy and has almost $101$k parameters.
We report validation performance, since the validation set is used to guide the design process of the NN.

\subsubsection{Neural Architecture Search}
\label{sec:nas}
There are many possible ways a NN architecture could look like. Usually, this is manually engineered by a domain expert. However, this process can be time consuming, especially when the NN should be applied to a novel domain (\eg new dataset for which there is no or little prior experience on which type of NN could work). 
NAS allows optimizing the architecture of a network in addition to the regular parameters, \ie it aims to find a good architecture automatically. NAS itself is a research field on its own; an overview can be found in \cite{NAS-Elsken2019}.
Moreover, hardware metrics can be included in the search, \eg the amount of memory or the number of operations, allowing architectures to be searched and optimized \wrt hardware considerations.
This is an important aspect for finding resource-efficient architectures that fit on an embedded device.
We substitute the manual design process by employing NAS.

\begin{figure}
	\centering
	\subfloat[Manually-found architecture]{\includegraphics[width=0.98\linewidth, keepaspectratio]{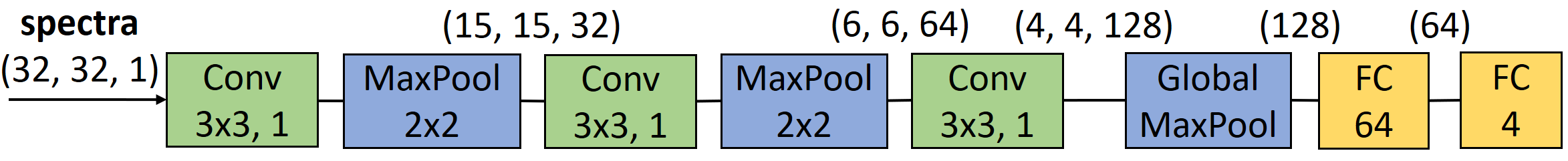}}
	\hfill
	\subfloat[Seed architecture for the NAS algorithm]{\includegraphics[width=0.70\linewidth, keepaspectratio]{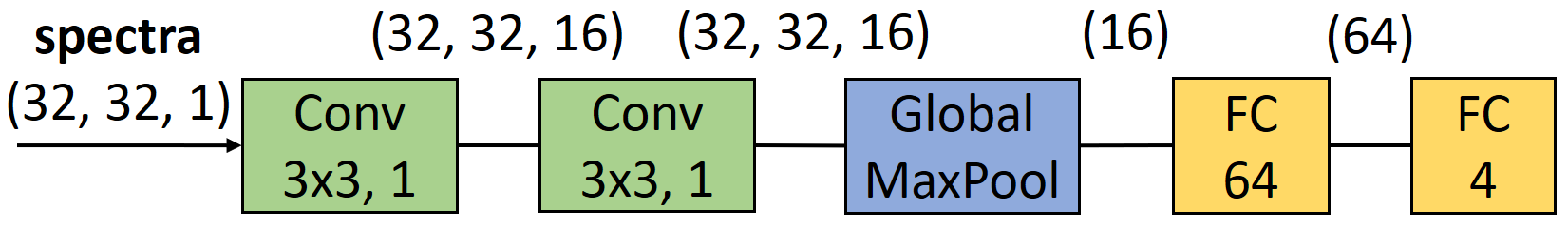}}
	\hfill
	\subfloat[DeepHybrid model built by attaching the reflection branch to an automatically-found architecture]{\includegraphics[width=0.88\linewidth, keepaspectratio]{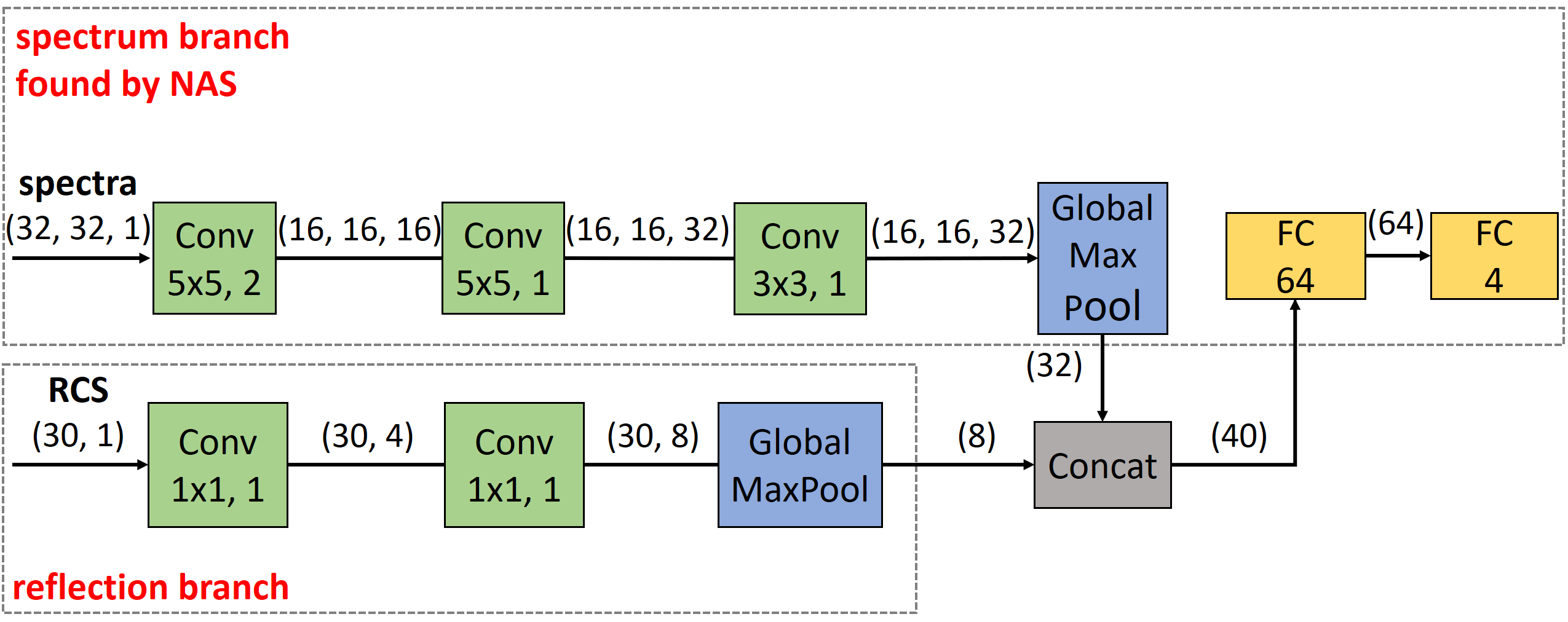}}
	\caption{Overview of the different neural network (NN) architectures: The NN from (a) was manually designed.
		(b) shows the NN from which the neural architecture search (NAS) method starts.
		Deploying the NAS algorithm yields a NN with similar accuracy, but with 7 times less parameters, depicted within the "found by NAS" box in (c). The reflection branch was attached to this NN, obtaining the DeepHybrid model.
		The layers are characterized by the following numbers.
		Convolutional (Conv) layer: kernel size, stride. Max-pooling (MaxPool): kernel size. Fully connected (FC): number of neurons.
		Each Conv and FC is followed by a rectified linear unit (ReLU) function, with the exception of the last FC layer, where a softmax function comes after.
		The numbers in round parentheses denote the output shape of the layer.} 
	\label{fig:arch}
\end{figure}

\begin{figure}[t]
	\centering
		\centering
	\subfloat[Pareto front: mean accuracy vs. number of parameters]{\includegraphics[width=0.75\linewidth, keepaspectratio]{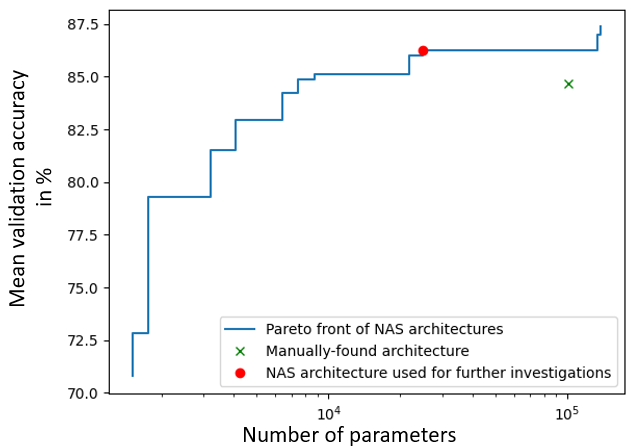}}
	\hfill
	\subfloat[Pareto front: mean accuracy vs. number of MACs]{\includegraphics[width=0.75\linewidth, keepaspectratio]{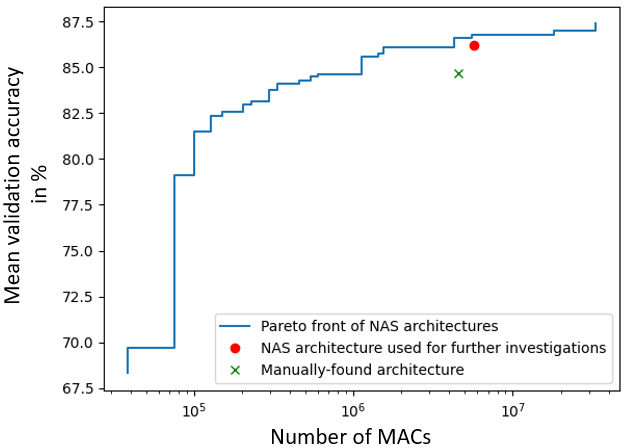}}
	\caption{Pareto fronts of architectures found by neural architecture search (NAS). (a) shows the Pareto front of mean validation accuracy vs. number of parameters. The red dot marks an automatically-found architecture with similar accuracy to the manually-found one from \ref{sec:manual} (green cross), but with almost 7 times less parameters. This is the architecture we further investigate in \ref{sec:hybrid} and \ref{sec:experiments}. (b) depicts the Pareto front of mean accuracy vs. number of MACs. The red dot is the same architecture as in (a). Note that there is a tradeoff among the $3$ optimization objectives of NAS, namely mean validation accuracy, number of parameters, and number of MACs. Thus, the architecture marked with the red dot is not optimal \wrt the number of MACs. 
}
	\label{fig:pareto_front}
\end{figure}

There are many search methods in the literature, each with advantages and shortcomings. 
Comparing search strategies is beyond the scope of this paper (\cf \cite{NAS-Elsken2019,real2019aging} for a detailed case study). 
Here, we chose to run an evolutionary algorithm \cite{deb2001multi} for obtaining a suitable NN to solve our task.
Evolutionary methods inherently enable exploration and parallelization \cite{stanley2019designing}, and can easily deal with multi-objective optimization tasks.
In our case, the search is governed by three simultaneous objectives: maximizing the mean validation accuracy, \ie \textit{performance}, minimizing the number of parameters, \ie \textit{memory efficiency}, and minimizing the number of multiply-accumulate operations (MACs), \ie small \textit{computational effort} during inference.
The search could lead to a network that fits on a radar sensor and runs in real time.
There is always a trade-off among the different objectives.
Therefore, in general, there does not exist one optimal solution but a set of optimal solutions, called Pareto-optimal solutions \cite{pareto-front}.
The goal of NAS is to find network architectures that are located near the true Pareto front.
We use a combination of the non-dominant sorting genetic algorithm II \cite{deb2002-NSGA2} and the regularized evolution algorithm \cite{real2019regularized}.
The search starts from a seed architecture, which is depicted in \figref{fig:arch} (b).
The following mutations to an architecture are allowed during the search: adding or removing convolutional (Conv) layers, adding or removing max-pooling layers, and changing the kernel size, stride, or the number of filters of a Conv layer.
Note that the manually-designed architecture depicted in \figref{fig:arch} (a) is included in the search space. 
During the NAS search, $1000$ different architectures were trained and evaluated. 
The search was conducted in a parallel setup, using 10 GPUs, and took around 2 hours to complete.
The Pareto fronts, mean accuracy vs. number of parameters and mean accuracy vs. number of MACs, obtained at the end of this automatic search are shown in \figref{fig:pareto_front}. 

For each architecture on the curve illustrated in \figref{fig:pareto_front} (a), the mean validation accuracy and the number of parameters were computed. The manually-designed NN is also depicted in the plot (green cross). It can be observed that NAS found architectures with similar accuracy, but with an order of magnitude less parameters. For further investigations, we pick a NN, marked with a red dot in \figref{fig:pareto_front} (a), with slightly better performance and approximately $7$ times less parameters than the manually-designed NN. Its architecture is presented in \figref{fig:arch} (c) as the sequence of layers within the "found by NAS" box.
Comparing the architectures of the automatically- and manually-found NN (see \figref{fig:arch} (a) and (c)), we can make the following observations. 
The NAS method prefers larger convolutional kernel sizes. 
Moreover, the automatically-found NN has a larger stride in the first Conv layer and does not contain max-pooling layers, \ie the input is downsampled only once in the network.
The automatically-found NN uses less filters in the Conv layers, which leads to less parameters than the manually-designed NN.

\figref{fig:pareto_front} (b) shows the Pareto front of mean accuracy vs. number of MACs, where the architecture marked with the red dot is the same as in \figref{fig:pareto_front} (a). This has a slightly better performance than the manually-designed one and a bit more MACs. Note that the red dot is not located exactly on the Pareto front. 
This shows that there is a tradeoff among the $3$ optimization objectives of NAS, \ie mean accuracy, number of parameters, and number of MACs. \figref{fig:pareto_front} (a) and (b) show only the tradeoffs between $2$ objectives. Therefore, the NN marked with the red dot is not optimal \wrt the number of MACs.
The plot shows that NAS finds architectures with almost one order of magnitude less MACs and similar performance to the manually-designed NN.

Unfortunately, there do not exist other DL baselines on radar spectra for this dataset. Using NAS, the accuracies of a lot of different architectures are computed. Therefore, comparing the manually-found NN with the NAS results is like comparing it to a lot of baselines at once. 

\subsection{DeepHybrid}
\label{sec:hybrid}
We build a hybrid model on top of the automatically-found NN (red dot in \figref{fig:pareto_front}) by attaching the reflection branch to it, see \figref{fig:arch} (c).
The reflection branch gets a $(30, 1)$ input that contains the radar cross-section (RCS) values corresponding to the reflections associated to the object to be classified. 
Each object can have a varying number of associated reflections. 
We choose a size of $30$ to ensure a fixed-size input, which is typically larger than the number of associated reflections, and set the remaining values to zero. 
This type of input can be interpreted as point cloud data \cite{pointnet}, therefore the design of this branch is inspired by \cite{pointnet}. 
The RCS input is processed by two convolutional layers with a $1\times1$ kernel, each followed by a rectified linear unit (ReLU) function. 
By design, these layers process each reflection in the input independently.
This is equivalent to a multi layer perceptron consisting of 2 layers with output shapes $(30,4)$ and $(30,8)$, where the parameters of each layer are shared across the first dimension, \ie the one which corresponds to the reflections.
This sequence ends with a global max-pooling layer, which ensures that the reflection branch is invariant \wrt the order of reflections in the input. The max-pooling operation is also a way to compute a global representation of all reflections. 

\subsection{Training Pipeline}
\label{sec:pipeline}
For all experiments presented in the following section, the NN is trained for $1000$ epochs, using the Adam optimizer \cite{kingma2017adam} with a learning rate of $0.003$ and batch size of $128$.
The training set is unbalanced, \ie the numbers of samples per class are different. In the considered dataset there are 11 times more car samples than two-wheeler or pedestrian samples, and 3 times more car samples than overridable samples. To overcome this imbalance, the loss function is weighted during training with class weights that are inversely proportional to the class occurrence in the training set.

\section{Experimental Results}
\label{sec:experiments}
\begin{figure}[t]
	\centering
	\subfloat[Spectrum branch model achieves a mean accuracy of $84.2\%$.]{\includegraphics[width=0.48\linewidth, keepaspectratio]{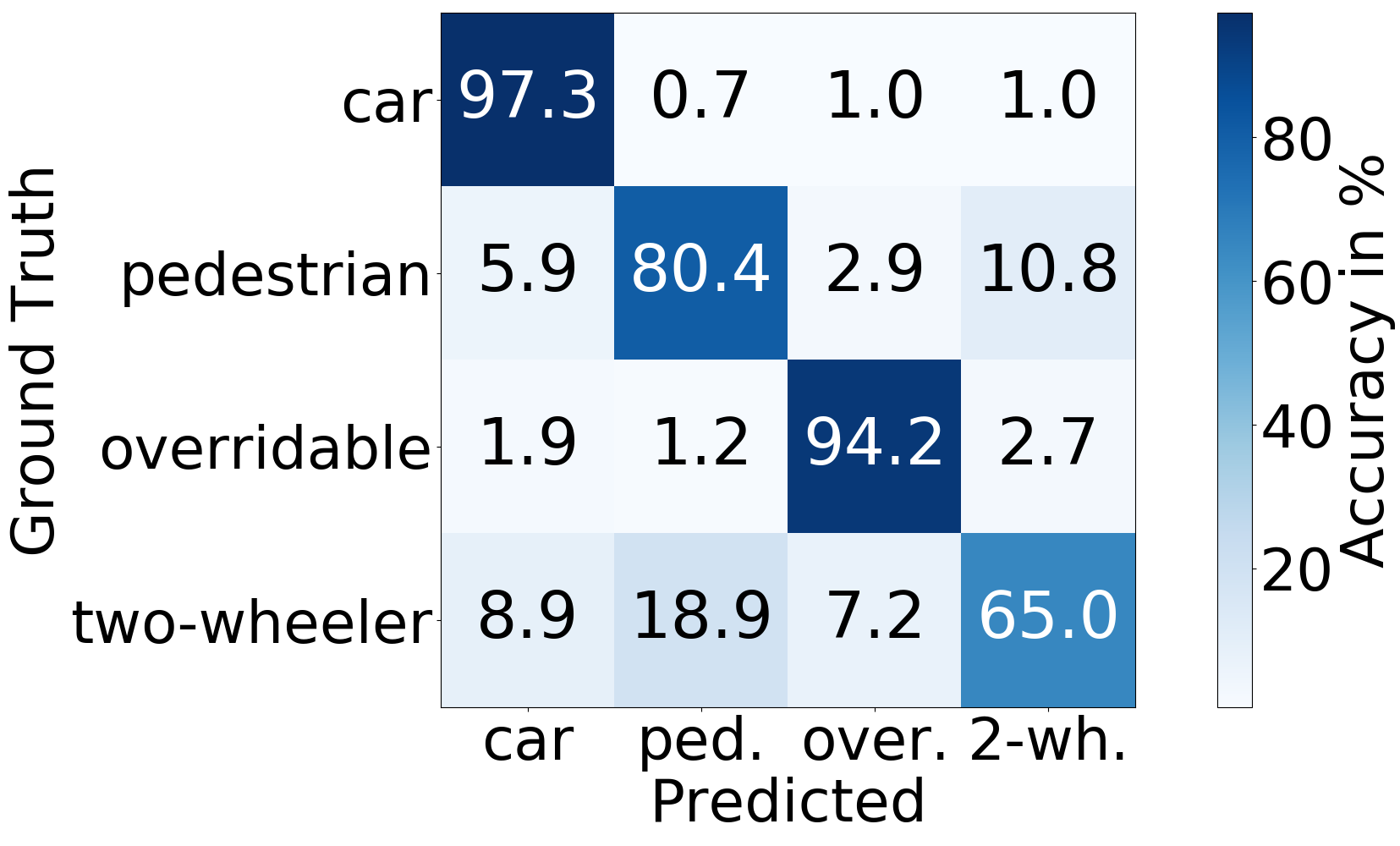}}
	\hfill
	\subfloat[DeepHybrid has a mean accuracy of $89.9\%$.]{\includegraphics[width=0.48\linewidth, keepaspectratio]{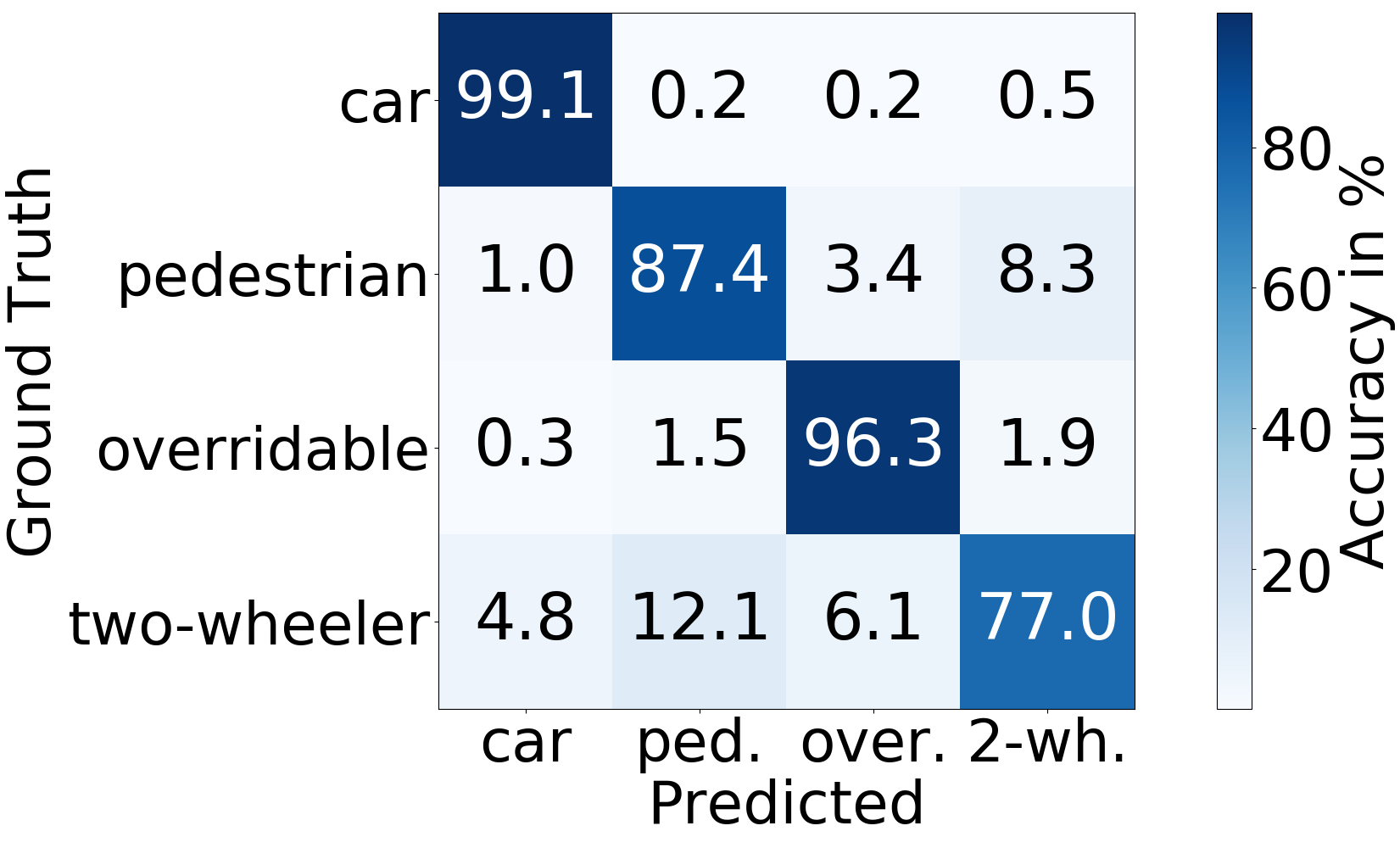}}
	\caption{Confusion matrices computed on the test set. The values in each row are divided by the corresponding number of class samples. Averaging the values on the main diagonal yields the reported mean accuracy.} 
	\label{fig:cnf_matrix}
\end{figure}
Each experiment is run 10 times using the same training and test set, but with different initializations for the NN's parameters.
The trained models are evaluated on the test set and the confusion matrices are computed.
A confusion matrix shows both the per class accuracies (\eg how well the model predicts a car sample as a car) and the confusions (\eg how often the model says a car sample is a pedestrian). 
The true classes correspond to the rows in the matrix and the columns represent the predicted classes. 
Each confusion matrix is normalized, \ie the values in a row are divided by the corresponding number of class samples.
In this way, we account for the class imbalance in the test set.
We report the mean over the 10 resulting confusion matrices.
For all considered experiments, the variance of the 10 confusion matrices is negligible, if not mentioned otherwise.

The confusion matrices of DeepHybrid introduced in \ref{sec:hybrid} and the spectrum branch model presented in \ref{sec:nas} are shown in \figref{fig:cnf_matrix}. It can be observed that using the RCS information in addition to the spectra helps DeepHybrid to better distinguish the classes. Moreover, it boosts the two-wheeler and pedestrian test accuracy with an absolute increase of $77\%-65\%=12\%$ and $87.4\%-80.4\%=7\%$, respectively. Nevertheless, both models mistake some pedestrian samples for two-wheeler, and vice versa.

The mean test accuracy is computed by averaging the values on the confusion matrix' main diagonal.
The spectrum branch model has a mean test accuracy of $84.2\%$, whereas DeepHybrid achieves $89.9\%$. 
For learning the RCS input, DeepHybrid needs $560$ parameters in addition to the already $25$k required by the spectrum branch. 
In conclusion, the RCS input yields an absolute improvement of $5.7\%$ in test performance at a cost of only about $2\%$ more parameters.
In comparison, the reflection branch model, \ie the reflection branch followed by the two FC layers, see \figref{fig:arch} (c), achieves $61.4\%$ mean test accuracy, with a significant variance of $10\%$. Such a model has $900$ parameters. Hence, the RCS information alone is not enough to accurately classify the object types. 

We also evaluate DeepHybrid against a classifier implementing the k-nearest neighbors (kNN) vote \cite{cunningham2020k}, in order to establish a baseline with respect to machine learning methods.
The kNN classifier predicts the class of a query sample by identifying its $k$ nearest neighbors. We use the L$^2$ distance. 
A majority vote over their classes yields the predicted class. 
In our implementation, the neighbors are weighted by the inverse of their distance to the query sample. 
We consider $k = 3, 4, 5, 7, 10$ and compute the corresponding confusion matrix on the validation set. 
The highest mean validation accuracy was obtained for $k = 4$, for which the mean test accuracy is $77.3\%$.
In comparison, DeepHybrid achieves $89.9\%$ mean test accuracy, so a $12.6\%$ absolute improvement when using a DL approach.

\section{Conclusion and Outlook}
How to best combine radar signal processing and DL methods to classify objects is still an open question.
This work demonstrates a possible solution: 1) A data preprocessing stage extracts sparse regions of interest (ROIs) from the radar spectra based on the detected and associated radar reflections. 2) A neural network (NN) uses the ROIs as input for classification.
This enables the classification of moving and stationary objects.
To improve classification accuracy, a hybrid DL model (DeepHybrid) is proposed, which processes radar reflection attributes and spectra jointly.
We showed that DeepHybrid outperforms the model that uses spectra only.
Manually finding a resource-efficient and high-performing NN can be very time consuming.
The paper illustrates that neural architecture search (NAS) algorithms can be used to automatically search for such a NN for radar data.
NAS finds a NN that performs similarly to the manually-designed one, but is 7 times smaller.
Future investigations will be extended by considering more complex real world datasets and including other reflection attributes in the NN's input.
The NAS algorithm can be adapted to search for the entire hybrid model.

\end{document}